\documentclass[10pt,twocolumn,letterpaper]{article}
\usepackage[accsupp]{axessibility}
\usepackage{iccv}
\usepackage{times}
\usepackage{epsfig}
\usepackage{graphicx}
\usepackage{amsmath}
\usepackage{amssymb}
\usepackage{multirow}
\usepackage{algorithm}
\usepackage{algpseudocode}
\usepackage{booktabs}
\usepackage{color}
\usepackage{xcolor}
\usepackage{cuted}
\usepackage{capt-of}
\usepackage[normalem]{ulem}
\usepackage{longtable}
\usepackage{multirow}
\usepackage{cite}
\newcommand{\comment}[1]{\textcolor{gray}{#1}}

\usepackage{xcolor}
\PassOptionsToPackage{hyphens}{url}\usepackage[pagebackref=true,breaklinks=true,letterpaper=true,colorlinks,
  citecolor=citecolor,bookmarks=false]{hyperref}
\definecolor{citecolor}{RGB}{34,139,34}

\def\iccvPaperID{9615}

\def\ours{InterDiff}

\iccvfinalcopy 
\ificcvfinal\fi

\begin{document}
\title{InterDiff: Generating 3D Human-Object Interactions with Physics-Informed Diffusion}

\author{Sirui Xu \quad  
Zhengyuan Li \quad
Yu-Xiong Wang$^*$ \quad
Liang-Yan Gui\thanks{Equal contribution.}\\
University of Illinois at Urbana-Champaign\\
{\tt\small\{siruixu2, zli138, yxw, lgui\}@illinois.edu}\\
{\tt\small\url{https://sirui-xu.github.io/InterDiff/}}}

\maketitle
\ificcvfinal\thispagestyle{empty}\fi

\begin{strip}\centering
\vspace{-6em}
\includegraphics[width=\textwidth]{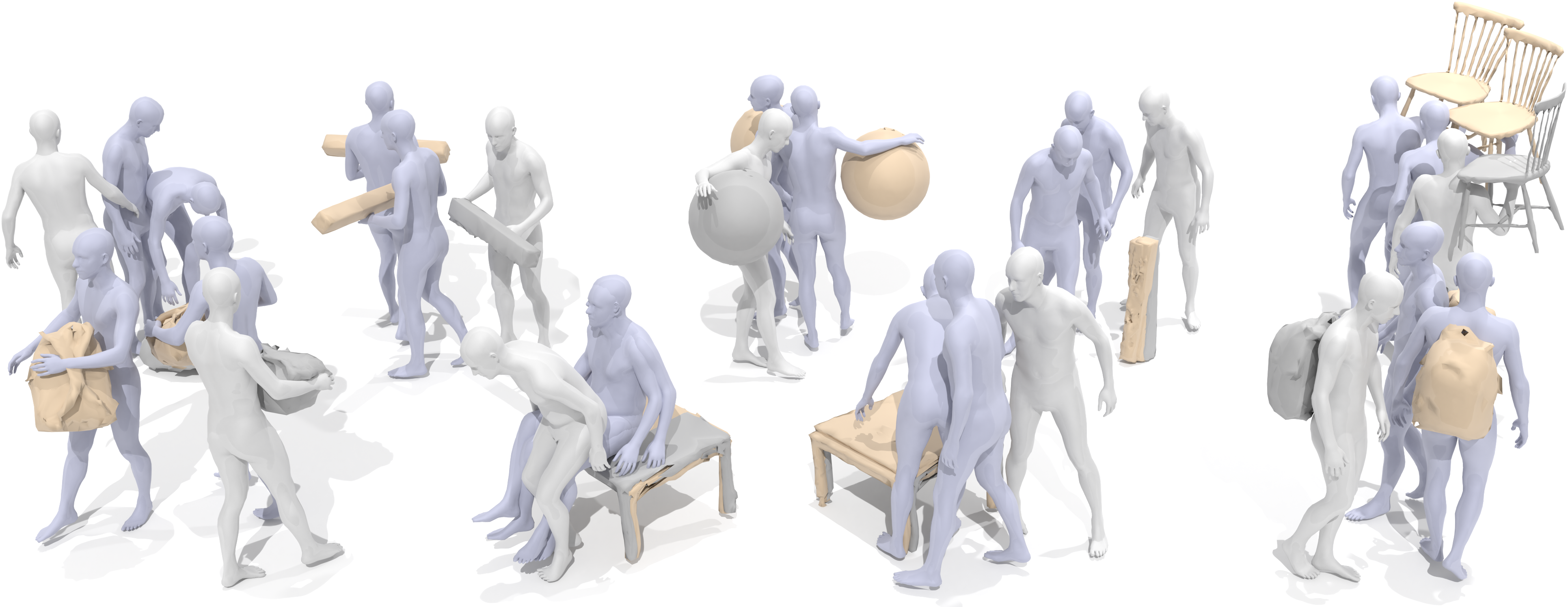}
\captionof{figure}{\textbf{A novel task of predicting 3D human-object interactions.} We provide 9 HOI sequences sampled every 40 frames at 30 FPS. Conditioned on past HOIs in gray meshes, our model generates long-term, diverse, and vivid HOIs, represented by the colored meshes.
\label{fig:teaser}}
\end{strip}

\begin{abstract}
This paper addresses a novel task of anticipating 3D human-object interactions (HOIs). Most existing research on HOI synthesis lacks comprehensive whole-body interactions with dynamic objects, e.g., often limited to manipulating small or static objects. Our task is significantly more challenging, as it requires modeling dynamic objects with various shapes, capturing whole-body motion, and ensuring physically valid interactions. To this end, we propose InterDiff, a framework comprising two key steps: (i) interaction diffusion, where we leverage a diffusion model to encode the distribution of future human-object interactions; (ii) interaction correction, where we introduce a physics-informed predictor to correct denoised HOIs in a diffusion step. Our key insight is to inject prior knowledge that the interactions under reference with respect to contact points follow a simple pattern and are easily predictable. Experiments on multiple human-object interaction datasets demonstrate the effectiveness of our method for this task, capable of producing realistic, vivid, and remarkably long-term 3D HOI predictions.
\end{abstract}
\section{Introduction}

Being able to ``look into the future'' is a remarkable cognitive hallmark of humans. Not only
can we anticipate how people will move or behave in the near future, but we can also forecast how our actions will interact with the ever-changing environment based on past information. An automated system that accurately forecasts 3D human-object interactions (HOIs) would have significant implications for various fields, such as robotics, animation, and computer vision. However, existing work on HOI synthesis does not adequately reflect the real-world complexity, \eg, examining hand-object interactions from an ego-centric view~\cite{liu2020forecasting,liu2022joint}, synthesizing interactions of grasping small objects~\cite{ghosh2022imos}, representing HOIs in simplified skeletons~\cite{corona2020context,9714029,razali2023action}, or overlooking object dynamics~\cite{taheri2022goal,wu2022saga,kulkarni2023nifty}.

To overcome such limitations, in this work, we reformulate the task of {\em human-object interaction prediction}, where we aim to model and forecast 3D mesh-based whole-body movements and object dynamics simultaneously, as shown in Figure~\ref{fig:teaser}. This task presents several unique real-world challenges. \textbf{(i) High Complexity}: it requires the modeling of both full-body and object dynamics, which is further complicated by the considerable variability in object shapes. \textbf{(ii) Physical Validity}: the predicted interaction should be {\em physically} plausible. Specifically, the human body should naturally conform to the surface of the object when in contact, while avoiding any penetration. 

\begin{figure}
    \centering
    \includegraphics[width=\columnwidth]{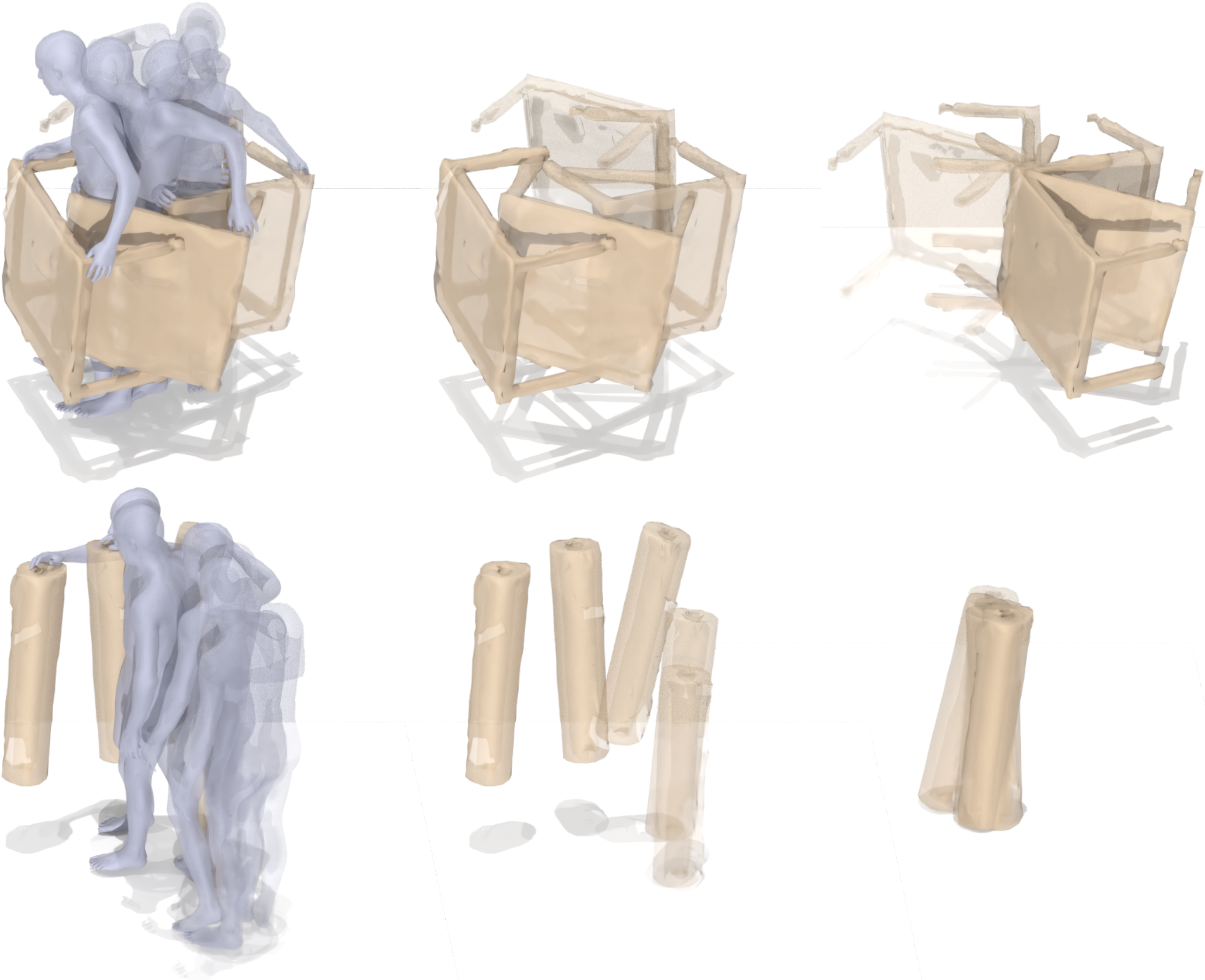}
    \caption{We present ground truth HOI sequences (\textbf{left}), object motions (\textbf{middle}), and objects relative to the contacts after coordinate transformations (\textbf{right}). Our key insight is to inject coordinate transformations into a diffusion model, as the relative motion shows simpler patterns that are easier to predict, \eg, rotating around a fixed axis (\textbf{top}) or being almost stationary (\textbf{bottom}).}
    \label{fig:teaser_2}
\end{figure}

A na\"ive approach would be to directly extend existing deep generative models that have been developed for human motion prediction, as exemplified by motion diffusion models~\cite{tevet2023human}, to capture the distribution of future human-object interactions. However, these models fail to incorporate the underlying physical laws that would ensure perceptually realistic predictions, thus introducing artifacts such as floating contact and penetration. This problem is amplified when autoregressive inference is utilized to synthesize long-term interactions, as errors accumulate over time. To this end, most existing research on 3D HOI synthesis~\cite{taheri2022goal,wu2022saga,ghosh2022imos} relies on post-hoc optimization to inject physical constraints. HOI synthesis has also been explored with simulators~\cite{liu2018learning,merel2020catch,hassan2023synthesizing,bae2023pmp} to ensure physical properties. Although plausible interactions can be generated, effort is required to build a physics simulation environment, \eg, registering objects with diverse shapes, as well as frictions, stiffness, and masses, which are hardly present in motion capture datasets. Moreover, considerable time is needed to train control policies to track realistic interactions.

Rather than relying on post-optimization or physics simulation, we introduce a pure learning-based method that utilizes a diffusion model with intuitive physics directly injected, which we call ``InterDiff.''
Our approach is based on the key observation that \textit{the short-term relative motion of an object with respect to the contact point follows a simple and nearly deterministic pattern}, despite the complexity of the overall interaction. For example, when juggling balls, their path reflects a complex pattern under the global coordinate system, due to the movement of the juggler. Yet, each ball simply moves up and down with respect to the juggler’s hand. We provide further illustrations of relative motion extracted from the BEHAVE dataset~\cite{bhatnagar22behave} in Figure~\ref{fig:teaser_2}.

Inspired by this, our InterDiff incorporates two components as follows. \textbf{(i) Interaction Diffusion}: a Denoising Diffusion Probabilistic Model (DDPM)-based generator~\cite{ho2020denoising} that models the distribution of future human-object interactions. \textbf{(ii) Interaction Correction}: a {\em novel physics-informed interaction predictor} that synthesizes the object's \textit{relative motion} with respect to regions in contact on the human body. We enhance this predictor by promoting simple motion patterns for objects and encouraging contact fitting of surfaces, which largely mitigates the artifacts of interactions produced by the diffusion model. By injecting the plausible dynamics back into the diffusion model iteratively, InterDiff generates vivid human motion sequences with realistic interactions for various 3D dynamic objects. {\em Another attractive property} of InterDiff is that its two components can be trained separately, while naturally conforming during inference without fine-tuning.

Our \textbf{contributions} are three-fold. \textbf{(i)} To the best of our knowledge, we are the {\em first} to tackle the task of mesh-based 3D HOI prediction. \textbf{(ii)} We propose the {\em first} diffusion framework that leverages past motion and shape information to generate future human-object interactions. \textbf{(iii)} We introduce a simple yet effective HOI corrector that incorporates physics priors and thus produces plausible interactions to infill the denoising generation. Extensive experiments validate the effectiveness of our framework, particularly for {\em out-of-distribution objects, and long-term autoregressive inference} where input past HOIs may be unseen in the training data. We attribute our improved generalizability to our important design strategies, such as the promotion of simple motion patterns and the anticipated interaction within a local reference system.

\section{Related Work}

\noindent{\bf Denoising Diffusion Models.}
Denoising diffusion models~\cite{sohl2015deep,song2020denoising,ho2020denoising,liu2022compositional} are equipped with a stochastic diffusion process that gradually introduces noise into a sample from the data distribution, following thermodynamic principles, and then generates denoised samples through a reverse iterative procedure. Recent work has extended them to the task of human motion generation~\cite{barquero2022belfusion,tevet2023human,zhang2022motiondiffuse,raab2023single,zhang2023t2m,yonatan2023,DiffPred,huang2023diffusion,chen2023humanmac,chen2023executing,karunratanakul2023gmd,dabral2023mofusion,zhang2023remodiffuse,wei2023understanding,sun2023towards,tian2023transfusion,zhang2023tedi}. For instance, MDM~\cite{tevet2023human} utilizes a transformer architecture to predict clean motion in the reverse process. We extend their framework to our HOI prediction task.
To generate conditional samples, a common strategy involves repeatedly injecting available information into the diffusion process. A similar idea applies to motion diffusion models~\cite{tevet2023human,yonatan2023,raab2023single} for motion infilling. Compared with PhysDiff~\cite{yuan2022physdiff}, which injects a motion imitation policy based on physics simulation into the diffusion process, we leverage a much simpler interaction correction step that is informed of the appropriate coordinate system to yield plausible interactions at a lower cost.

\noindent{\bf Human-Object Interaction.}
Despite recent advancements in human-object interaction learning, existing research has primarily focused on HOI detection~\cite{gkioxari2018detecting,ji2021detecting,wu2022mining,zhou2022human,xie2023visibility,chen2023detecting,zhu2023diagnosing}, reconstruction~\cite{xie2022chore,zhang2020perceiving,wang2022reconstructing,petrov2023object,hou2023compositional,kim2023ncho}, and generating humans that interact with static scenes~\cite{cao2020long,hassan2021populating,wang2021synthesizing,wang2021scene,wang2022towards,wang2022humanise,huang2023diffusion,zhao2022compositional,Zhao:ICCV:2023,tendulkar2022flex,zhang2023roam}. Most attempts have been made to synthesize only hand-object interactions in computer graphics~\cite{li2007data,pollard2005physically,zhang2021manipnet}, computer vision~\cite{kry2006interaction,karunratanakul2020grasping,taheri2020grab,corona2020ganhand,jiang2021hand,grady2021contactopt,li2023task,ye2023affordance,zheng2023cams,zhou2022toch}, and robotics~\cite{brahmbhatt2019contactgrasp,detry2010refining,hsiao2006imitation,li2007data}. 
Generating whole-body interactions, such as approaching and manipulating static~\cite{taheri2022goal,wu2022saga,kulkarni2023nifty,zhang2022couch}, articulated~\cite{lee2023locomotion,xu2021d3dhoi}, and dynamic objects~\cite{ghosh2022imos} has also been a growing topic.
The task of synthesizing humans interacting with dynamic objects has been explored based on first-person vision~\cite{liu2020forecasting,liu2022joint} and skeletal representations~\cite{corona2020context,9714029,razali2023action} on skeleton-based datasets~\cite{Mandery2015a,Mandery2016b,krebs2021kit}. 
In humanoid control, progress on full-body HOI synthesis has been made with kinematic-based approaches~\cite{starke2019neural,starke2020local} and in the application of physics simulation environments~\cite{liu2018learning,chao2021learning,merel2020catch,hassan2023synthesizing,bae2023pmp,yang2022learning,xie2022learning,xie2023hierarchical,pan2023synthesizing}.
However, most approaches have limitations regarding action and object variation, such as focusing on approaching or manipulating objects on \eg, the GRAB~\cite{taheri2020grab} dataset. 
Recent datasets~\cite{bhatnagar22behave,jiang2022chairs,huang2022intercap,zhang2023neuraldome,fan2023arctic} are established to address the above limitations and provide 3D interactions with richer objects and actions, setting the stage for achieving our task.

\noindent{\bf Human Motion and Object Dynamics.}
Generative modeling, including variational autoencoders (VAEs)~\cite{autoencoder}, generative adversarial networks~\cite{GAN}, normalizing flows~\cite{JimenezRezende2015VariationalIW}, and diffusion models~\cite{sohl2015deep,song2020denoising}, has witnessed significant progress recently, leading to attempts for skeleton-based human motion prediction~\cite{apratim18cvpr2,Dilokthanakul2016DeepUC,Gurumurthy_2017_CVPR,yuan2019diverse,yuan2020dlow,mao2021generating,xu22stars,barquero2022belfusion}. Moreover, research has expanded beyond skeleton generation and utilized statistical models such as SMPL~\cite{loper2015smpl} to generate 3D body animations~\cite{zhang2021mojo,mao2022weakly,petrovich22temos,petrovich2021action,hassan_samp_2021,tevet2022motionclip,zhang2022wanderings,xu2023stochastic}. Our study employs SMPL parameters to drive the 3D mesh of the human body on the BEHAVE dataset~\cite{bhatnagar22behave}, while also extending the method to skeleton-based datasets~\cite{9714029}, demonstrating its broad applicability.
Predicting object dynamics has also received increasing attention~\cite{mrowca2018flexible,RempeDynamics2020,driess2022learning,ye2019compositional,zhu2018object}. Different from solely predicting the human motion or the object dynamics, our method jointly models their interactions. 
\section{Methodology}\label{method}
\begin{figure*}
    \centering
    \includegraphics[width=\textwidth]{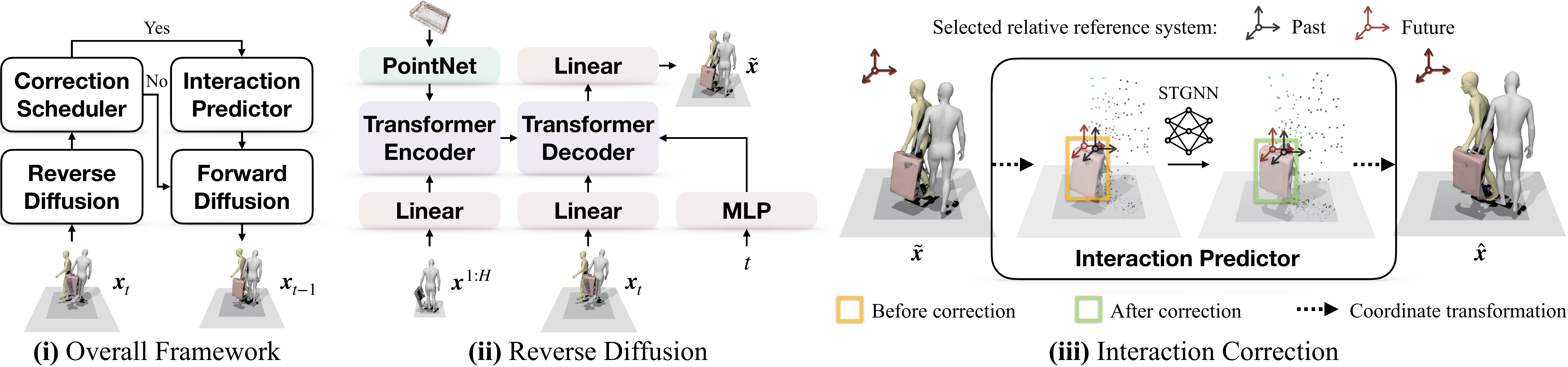}
    \caption{\textbf{Overview of \ours.}
    \textbf{(i)} We combine a Correction Scheduler and an Interaction Predictor with the diffusion framework to correct a denoised HOI. The Correction Scheduler determines whether the current denoised HOI needs correction. If so, we fuse the additional prediction generated by the Interaction Predictor into the denoised HOI. \textbf{(ii)} Our reverse diffusion employs a transformer architecture conditioned on the encoded object shape and the past HOI. \textbf{(iii)} We transform object motion under the reference system selected by the Correction Scheduler, predict future motion via STGNN, and transform it back to the ground system. Markers are in point clouds.}
    \label{fig:method}
\end{figure*}

\noindent{\bf Problem Formulation: Human-Object Interaction Prediction.} We denote a 3D HOI sequence with $H$ historical frames and $F$ future frames as $\boldsymbol x=[\boldsymbol x^1, \boldsymbol x^2,\ldots, \boldsymbol x^{H+F}]$,
where $\boldsymbol x^i$ consists of human pose state $\boldsymbol h^i$ and object pose state $\boldsymbol o^i$. Human pose state $\boldsymbol h^i \in \mathbb{R}^{J \times D_h}$ is defined by $J$ joints with a $D_h$-dimensional representation at each joint, which can be joint position, rotation, velocity, or their combination. Object pose state $\boldsymbol o^i$ has $D_o$ features, including \eg, the position of the center, and the rotation of the object \wrt the template. Note that the specific meanings of these states are dataset-dependent and will be explained in detail in Sec.~\ref{experiments}.
Given object shape information $\boldsymbol c$, our goal is to predict a 3D HOI sequence $\boldsymbol x_0$ that is (i) close to the ground truth $\boldsymbol x$ in future $F$ frames, and (ii) physically valid.

\noindent{\bf Overview.} As shown in Figure~\ref{fig:method}, InterDiff consists of interaction diffusion and correction.
In Sec.~\ref{diffusion}, we introduce interaction diffusion, which includes the forward and reverse diffusion processes. We explain how we extract shape information for the diffusion model. We then detail interaction correction in Sec.~\ref{correction}, including correction schedule and interaction prediction steps. Our {\em key insight} is applying interaction correction to implausible denoised HOIs. Given a denoised HOI after each reverse diffusion process, the correction scheduler determines if this denoised HOI needs correction, and infers a {\em reference system} based on contact information extracted from this intermediate result (Sec.~\ref{schedule}). If the correction is needed, we pass the denoised HOI and the inferred reference system to an interaction predictor, which forecasts plausible object motion under the identified reference. Afterward, we inject this plausible motion back into the denoised HOI for further denoising iterations (Sec.~\ref{prediction}). Notably, interaction diffusion and correction do not need to be coupled {\em during training}. Instead, they can be composed during inference \textit{without fine-tuning}.

\subsection{Interaction Diffusion}\label{diffusion}

\noindent{\bf Basic Diffusion Model.} Our approach incorporates a diffusion model, generating samples from isotropic Gaussian noise by iteratively removing the noise at each step. More specifically, to model a distribution $\boldsymbol x_0 \sim q(\boldsymbol x_0)$, the forward diffusion process follows a Markov chain of T steps, giving rise to a series of time-dependent distributions $q(\boldsymbol x_t|\boldsymbol x_{t-1})$. These distributions are generated by gradually injecting noise into the samples until the distribution of $\boldsymbol x_T$ is close to $\mathcal{N}(\boldsymbol 0, \mathbf I)$. Formally, this process is denoted as
\begin{align}\label{eq:1}
\begin{split}
    q(\boldsymbol x_1,\ldots,\boldsymbol x_T|\boldsymbol x_0) &= \prod_{t=1}^Tq(\boldsymbol x_t|\boldsymbol x_{t-1})\\
    q(\boldsymbol x_t|\boldsymbol x_{t-1}) &= \mathcal{N}(\sqrt{\beta_t}\boldsymbol x_{t-1} + (1 - \beta_t)\mathbf I),
\end{split}
\end{align}
where $\beta_t \in (0, 1)$ is the variance of the Gaussian noise injected at time $t$, and we define $\beta_0 = 0$.

Here, we adopt the Denoising Diffusion Probabilistic Model (DDPM)~\cite{ho2020denoising} for motion prediction, given that it can sample $\boldsymbol x_t$ directly from $\boldsymbol x_0$ without intermediate steps:
\begin{align}\label{eq:2}
\begin{split}
    q(\boldsymbol x_t|\boldsymbol x_0) &= \mathcal{N}(\sqrt{\bar \alpha_t}\boldsymbol x_{0} + (1 - \bar \alpha_t)\mathbf I)\\
    \boldsymbol x_t &= \sqrt{\bar \alpha_t}\boldsymbol x_{0} + \sqrt{1 - \bar \alpha_t}\boldsymbol \epsilon,
\end{split}
\end{align}
where $\alpha_t = 1 - \beta_t, \bar \alpha_t = \prod_{t'=0}^t\alpha_{t'}$, and $\boldsymbol \epsilon \sim \mathcal{N}(\boldsymbol 0, \mathbf I)$.

The reverse process of diffusion gradually cleans $\boldsymbol x_T \sim \mathcal{N}(\boldsymbol 0, \mathbf I)$ back to $\boldsymbol x_0$. Following~\cite{ramesh2022hierarchical,tevet2023human, raab2023single}, we directly recover the clean signal $\Tilde{\boldsymbol x}$ at each step, instead of predicting the noise $\boldsymbol \epsilon$~\cite{ho2020denoising} that was added to $\boldsymbol x_0$ in Eq.~\ref{eq:2}. This iterative process at step $t$ is formulated as
\begin{align}\label{eq:0}
\begin{split}
    \Tilde{\boldsymbol x} &= \mathcal{G}(\boldsymbol x_t, t, \boldsymbol c)\\
    \boldsymbol x_{t-1} &= \sqrt{\bar \alpha_{t-1}}\Tilde{\boldsymbol x} + \sqrt{1 - \bar \alpha_{t-1}}\boldsymbol \epsilon,
\end{split}
\end{align}
where $\mathcal{G}$ is a network estimating $\Tilde{\boldsymbol x}$ given the noised signal $\boldsymbol x_t$ and the condition $\boldsymbol c$ at step $t$, and $\boldsymbol \epsilon \sim \mathcal{N}(\boldsymbol 0, \mathbf I)$.

\noindent{\bf Interaction Diffusion Model.} While most existing \textit{human motion diffusion models}~\cite{raab2023single,tevet2023human,yuan2022physdiff,zhang2022motiondiffuse} can predict future frames by infilling ground truth past motion into the denoised motion at each diffusion step, we observe that encoding the historical motion $\boldsymbol x^{1:H}$ as a condition leads to better performance in our task. A similar design is used in~\cite{barquero2022belfusion,wei2023human} but for conventional human motion prediction.
Now our model $\mathcal{G}$ includes a transformer encoder that encodes $\boldsymbol x^{1:H}$ along with the object shape embedding $\boldsymbol c$ from a PointNet~\cite{qi2017pointnet}, shown in Figure~\ref{fig:method}(b). The input denoised HOI $\boldsymbol x_t$ at time step $t$ is linearly projected into the transformer combined with a standard positional embedding. A feed-forward network also maps the time step $t$ to the same dimension. The decoder then maps these representations to the estimated clean HOI $\Tilde{\boldsymbol x}$. $\mathcal{G}$ is optimized by the objective:
\begin{align}\label{eq:3}
\begin{split}
    \mathcal{L}_r = \mathbb{E}_{t \sim [1, T]} \|\mathcal{G}(\boldsymbol x_t, t, \boldsymbol c) - \boldsymbol x\|^2_2.
\end{split}
\end{align}
We further disentangle this objective into rotation and translation losses for both human state $\boldsymbol h$ and object state $\boldsymbol o$, and re-weight these losses. We also introduce velocity regularizations, as detailed in the Supplementary.

\subsection{Interaction Correction}\label{correction}
Given that deep networks do not inherently model fundamental physical laws, generating plausible interactions even for state-of-the-art generative models trained on large-scale datasets can be challenging. Instead of relying on post-hoc optimization or physics simulation to promote physical validity, we {\em embed an interaction correction step within the diffusion framework}. This is motivated by the fact that the diffusion model produces intermediate HOI $\tilde{\boldsymbol x}$ at each diffusion step, allowing us to blend plausible dynamics into implausible regions and still generate seamless results. Remarkably, we achieve such plausible interactions with a simple {\em physics-informed} correction step. This is greatly attributed to the essential inductive bias induced -- \eg, even though human and object motion can be complicated, the relative object motion in an appropriate reference system follows a simple pattern that is easier to predict.

\begin{algorithm}
    \caption{\ours: given a diffusion model $\mathcal{G}$, a correction scheduler $\mathcal{S}$, an interaction predictor $\mathcal{P}$, hyperparameters $\epsilon_1, \epsilon_2, \{\bar \alpha_{t}\}_{t=1}^T$}
    \label{algo:Interdiff}
    \begin{algorithmic}[1]
    \State \textbf{Input}: condition $\boldsymbol{c}$
    \State \textbf{Output}: the clean HOI $\boldsymbol{x}_0$ with correction
    \State $\boldsymbol{x}_T \sim \mathcal{N}(\boldsymbol 0, \mathbf I)$
    \For{$t$ from $T$ to 0}
    \State \comment{\# Reverse Diffusion}
    \State $\Tilde{\boldsymbol x} \leftarrow \mathcal{G}(\boldsymbol x_t, t, \boldsymbol c)$
    \State \comment{\# Correction Schedule}
    \State Obtain the contact and penetration states $\boldsymbol C, \boldsymbol P$ 
    \If{$\mathcal{S}(\boldsymbol P, \epsilon_1, \boldsymbol C, \epsilon_2, t)$}
        \State Obtain the reference system $s$
        \State \comment{\# Interaction Prediction}
        \State $\hat{\boldsymbol x} \leftarrow \mathcal{P}(\Tilde{\boldsymbol x}, s)$
        \State \comment{\# Interaction Blending}
        \State $\Tilde{\boldsymbol x} \leftarrow \Tilde{\boldsymbol x} \times \frac{t}{T} + \hat{\boldsymbol x} \times (1 - \frac{t}{T})$
    \EndIf
    \State \comment{\# Forward Diffusion}
    \State $\boldsymbol{x}_{t-1} \sim \mathcal{N}(\sqrt{\bar \alpha_{t-1}}\tilde{\boldsymbol x}, (1 - \bar \alpha_{t-1}) \mathbf{I})$
    \EndFor
    \State \textbf{return} ${\boldsymbol x_0}$
    \end{algorithmic}
\end{algorithm}

\subsubsection{Correction Schedule} \label{schedule}
Similar to PhysDiff~\cite{yuan2022physdiff}, we only consider performing corrections every few diffusion steps in late iterations, as early denoising iterations primarily produce noise with limited information. 

For 3D HOIs represented by meshes, we set additional constraints based on geometric clues to determine the steps to apply corrections. As demonstrated in Algorithm~\ref{algo:Interdiff}, given \textit{the current denoised HOI} $\tilde{\boldsymbol x}$, we first obtain the contact and penetration states in the future $F$ frames. Let $\boldsymbol v_h \in \mathbb{R}^{F \times V_h \times 3}$ be the human vertices where $V_h$ is the number of vertices, and $\mathbf{sdf}$ be a series of human body's signed distance fields~\cite{park2019deepsdf} in the future $F$ frames.
Specifically, for the SMPL representations~\cite{loper2015smpl}, the vertices and $\mathbf{sdf}$ can be derived from the skinning function, using the body shape and pose parameters in $\tilde{\boldsymbol x}$ as input. We can also obtain the future sequence of object point clouds $\boldsymbol v_o \in \mathbb{R}^{F \times V_o \times 3}$ from the object state in the denoised HOI $\tilde{\boldsymbol x}$, where $V_o$ is the number of object vertices. Based on the distance measurement, the contact state $\boldsymbol C \in \mathbb{R}^{F \times V_h}$ and the penetration state $\boldsymbol P \in \mathbb{R}^{F}$ are defined as,
\begin{align}\label{eq:4}
\begin{split}
    \boldsymbol C^i[j] &= \min_{k=1,\ldots,V_o} \|\boldsymbol v_h^i[j]-\boldsymbol v_o^i[k]\|_2, \ j=1,\ldots,V_h\\
    \boldsymbol P^i &= \sum_{k=1,\ldots,V_o}-\min\{\mathbf{sdf}(\boldsymbol v_o^i[k]), 0\},
\end{split}
\end{align}
where $\boldsymbol v_h^i[j] \in \mathbb{R}^{3}$, $\boldsymbol v^i_o[k] \in \mathbb{R}^{3}$ are $j$-th and $k$-th vertex on human and object at frame $i \in \{ H+1, \ldots, H+F\}$, respectively.

The correction scheduler $\mathcal{S}$ serves two main functions. One is to determine whether the current denoised HOI $\tilde{\boldsymbol x}$ requires correction. 
Guided by the contact information from $\tilde{\boldsymbol x}$, 
we only perform correction when the diffusion model is likely to make a mistake -- (i) \textit{penetration already exists}, defined as $\|\boldsymbol P\| > \epsilon_1$; or (ii) \textit{no contact happens}, defined as $\min_j\|\boldsymbol C[j] \| > \epsilon_2$. $\epsilon_1$ and $\epsilon_2$ are two hyperparameters. 
We only apply these constraints to mesh-represented HOIs, as the contact in skeletal HOIs is ill-defined.

The second function is to decide which reference system to use in Sec~\ref{prediction}. We define a set of markers $\mathcal{M}$~\cite{zhang2021mojo} to index $67$ human vertices as potential reference points for efficiency, instead of using all the $V_h$ vertices. We operate on the contact state $\boldsymbol{C}$ and get the index of the reference system $s$, as follows:
\begin{align}
\begin{split}
    s &= \left\{
    \begin{array}{cc}
    -1, & \mbox{if } \min\limits_{j \in \mathcal{M}} \|\boldsymbol C[j] \| \geq \epsilon_2 \\
    \underset{j \in \mathcal{M}}{\arg\min} \|\boldsymbol C[j] \|, & \mbox{o.w.}
    \end{array}
    \right.
\end{split}
\end{align}
This selection process means that we retain the default ground reference system if there is no contact; otherwise, we determine the reference point as the marker on the human body surface that is in contact with the object.

For skeletal HOIs, we follow a similar way to define the contact state $\boldsymbol C \in \mathbb{R}^{F \times J_h}$ for the purpose of capturing the reference system, despite its ill-posedness, as follows:
\begin{equation}\label{eq:skel}
    \boldsymbol C^i[j] = \min_{k=1,\ldots,J_o}\|\boldsymbol j_h^i[j]-\boldsymbol j_o^i[k]\|_2, \ j=1,\ldots,J_h\\,
\end{equation}
where $\boldsymbol j_h \in \mathbb{R}^{F \times J_h \times 3}$ represents $J_h$ human joints and $\boldsymbol j_o \in \mathbb{R}^{F \times J_o \times 3}$ represents $J_o$ object keypoints. We define the reference system $s$ based on joints rather than markers:
\begin{align}
\begin{split}
    s &= \left\{
    \begin{array}{cc}
    -1, & \mbox{if } \min\limits_{j=1,\ldots,J_h} \|\boldsymbol C[j] \| \geq \epsilon_2 \\
    \underset{j=1,\ldots,J_h}{\arg\min} \|\boldsymbol C[j] \|, & \mbox{o.w.}
    \end{array}
    \right.
\end{split}
\end{align}

\subsubsection{Interaction Prediction} \label{prediction}

Given the past object motion and the trajectories of human markers/joints in both \textit{past and future}, we now predict future object motions under different references. We first apply coordinate transformations to the past object motion (which is by default under the ground reference system) and obtain relative motions with respect to {\em all} markers/joints. Then we formulate the object motions, either under the ground reference system or relative to each marker/joint, {\em collectively} as a spatial-temporal graph $\boldsymbol{G}^{1:H}$. For example, given $|\mathcal{M}|$ markers, we define $\boldsymbol{G}^{1:H} \in \mathbb{R}^{H\times (1 + |\mathcal{M}|) \times D_o}$, where $D_o$ is the number of features for object poses as defined previously. Here, $1 + |\mathcal{M}|$ correspond to $1$ ground reference system and $|\mathcal{M}|$ marker-based reference systems.

Given the spatial-temporal graph $\boldsymbol{G}^{1:H}$, we use a spatial-temporal graph neural network (STGNN)~\cite{xu22stars} to process the past motion graph $\boldsymbol{G}^{1:H}$ and obtain $\boldsymbol{G}^{H:H+F}$ that represents future object motions in these systems.
Then, we acquire a specific object relative motion $\boldsymbol{G}^{H:H+F}[s+1]$, under the reference system $s$ specified in Sec.~\ref{schedule}. We transform this predicted reference motion back to the ground system. The resulting object motion is defined as $\hat{\boldsymbol x} = \mathcal{P}(\Tilde{\boldsymbol x}, s)$, where the interaction predictor $\mathcal{P}$ performs the above operations. We blend the original denoised HOI $\tilde{\boldsymbol x}$ with this newly obtained HOI $\hat{\boldsymbol x}$, denoted as $\Tilde{\boldsymbol x} \times \frac{t}{T} + \hat{\boldsymbol x} \times (1 - \frac{t}{T})$. 

\begin{figure*}
    \centering
    \includegraphics[width=\textwidth]{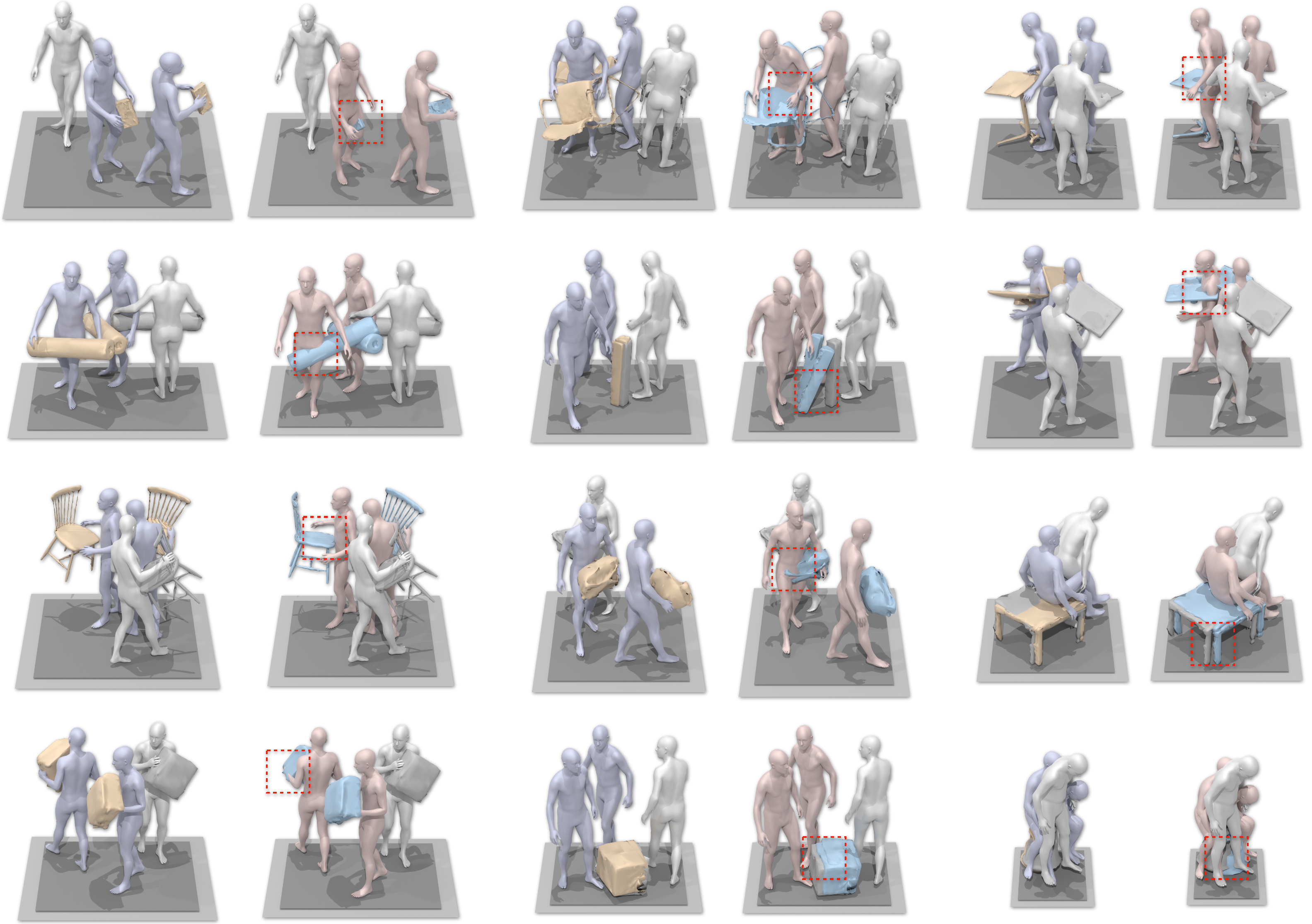}
    \caption{\textbf{Qualitative comparisons} on the BEHAVE dataset~\cite{bhatnagar22behave}. We show starting HOIs in gray and predicted HOIs sampled every 40 frames (30 FPS). The blue and red human meshes denote the results from \ours~with and without interaction correction, respectively. The injected correction step helps mitigate contact floating and penetration artifacts, and maintain static objects when there is no contact.}
    \vspace{-0.2em}
    \label{fig:BEHAVE_qual}
\end{figure*}

Informed by the reference system, we argue that the motion after coordinate transformation follows a simpler pattern and becomes easier for the network to predict and maintain physical validity. To further promote the simple motion pattern, in this STGNN, we use DCT/IDCT~\cite{ahmed1974discrete} as a preprocessing step in accordance with~\cite{mao2019learning}. We also find that a small number of frequency bases work well for predicting relative object motion. To decouple STGNN training from diffusion, we directly use \textit{clean HOI} data for training and perform inference on \textit{denoised HOIs}. We introduce learning objectives to promote contact and penalize penetration, which are detailed in the Supplementary.
\begin{figure}
    \centering
    \includegraphics[width=\columnwidth]{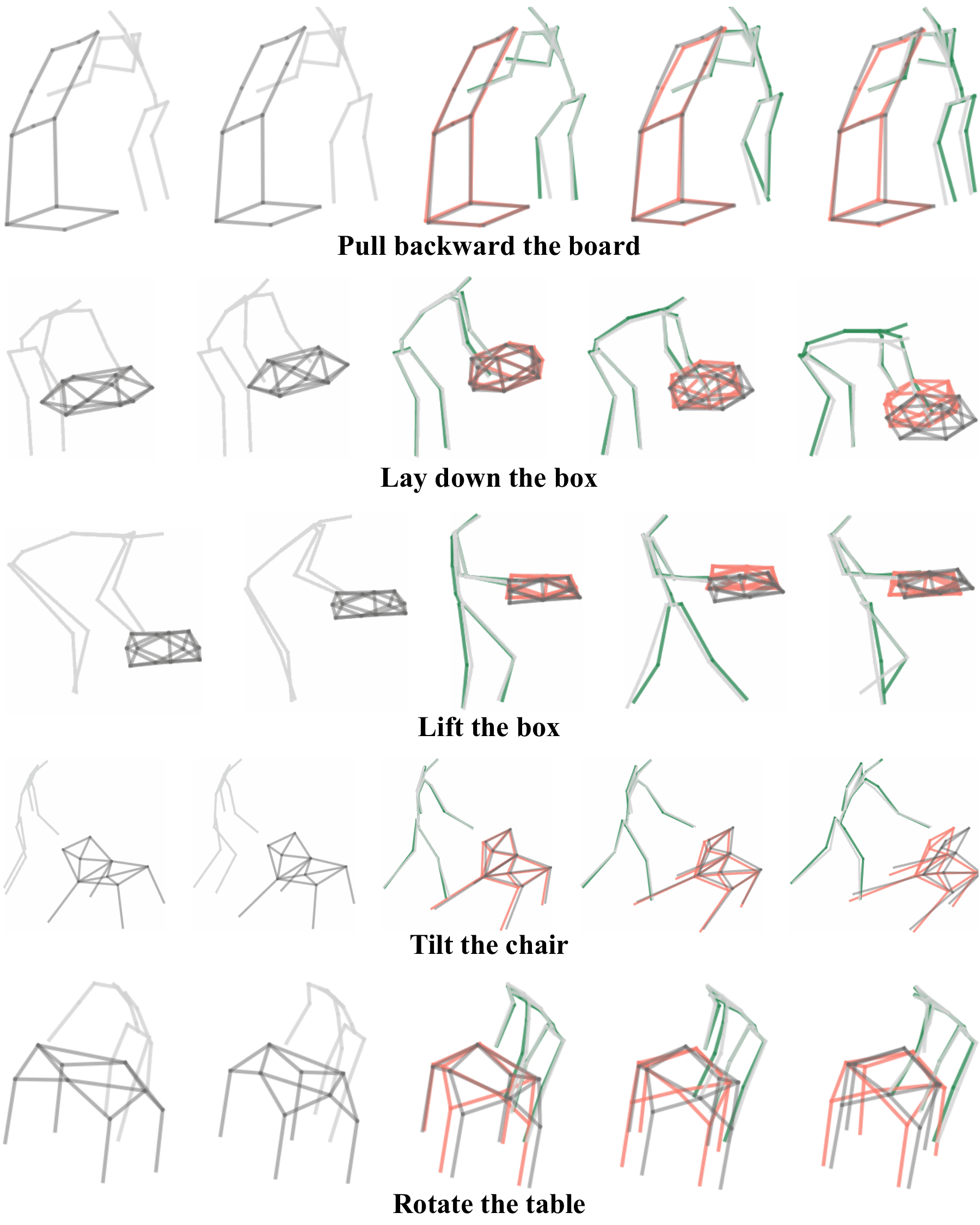}
    \caption{\textbf{Qualitative results} on interactions with \textit{unseen objects} on the Human-Object Interaction dataset~\cite{9714029}. The predicted skeletons and objects are green and red respectively while GT is gray. We show five frames at 0.4, 0.8, 1.2, 1.6, and 2.0s.}
    \label{fig:HOI_qual}
\end{figure}
\section{Experiments}\label{experiments}

\subsection{Experimental Setup}
\noindent{\bf Datasets.}
We conduct the evaluation on three datasets pertaining to 3D human-object interaction. BEHAVE~\cite{bhatnagar22behave} encompasses recordings of 8 individuals engaging with 20 ordinary objects at a framerate of 30 Hz. SMPL-H~\cite{loper2015smpl,MANO} is used to represent the human, and we represent the object pose in 6D rotation~\cite{zhou2019continuity} and translation. We adhere to the official train and test split originally proposed in HOI detection~\cite{bhatnagar22behave}. During training, our model forecasts 25 future frames after being provided with 10 past frames, and we can generate longer motion sequences autoregressively during inference. GRAB~\cite{taheri2020grab} is a dataset that records whole-body humans grasping small objects, including 1,334 videos, which we downsample to 30 FPS. We investigate the {\em cross-dataset generalizability} -- we train our method on the BEHAVE dataset and test it on the GRAB dataset. 
The Human-Object Interaction dataset~\cite{9714029} comprises 6 individuals and 384,000 frames recorded at a framerate of 240 Hz. We follow the official data preprocessing guidelines and extract the sequences at a framerate of 10 Hz. In total, 18,352 interactive sequences with a length of 20 frames are obtained, and 582 of these sequences include objects that are not seen during training and are directly used for evaluation. Following~\cite{9714029}, our model is trained to forecast 10 future frames given 10 past frames. Unlike the above two datasets, we employ a 21-joint skeleton to represent the human pose, and 12 key points for objects.

\noindent{\bf Metrics.}
Based on the established evaluation metrics in the literature~\cite{Kaufmann_2021_ICCV,9714029,8444061,hinterstoisser2013model}, we introduce a set of metrics to evaluate this new task as follows.
(i) \textbf{MPJPE-H}: the average $l_2$ distance between the predicted and ground truth joint positions in the global space. For SMPL-represented HOIs, joint positions can be obtained through forward kinematics.
(ii) \textbf{Trans.~Err.}: the average $l_2$ distance between the predicted and ground truth object translations. (iii) \textbf{Rot.~Err.}: the average $l_1$ distance between the predicted and ground truth quaternions of the object. (iv) \textbf{MPJPE-O}: the average $l_2$ distance between the predicted and ground truth object key points in the global space. This is only reported for the Human-Object Interaction dataset, where the object is abstracted into key points. (v) \textbf{Pene.}: the average percentage of object vertices with non-negative human signed distance function~\cite{park2019deepsdf} values.
Note that (i)(ii)(iv) are in $mm$, (iii) is in $10^{-3}$ radius, and (v) is in $10^{-2}\%$. In Table~\ref{table:BEHAVE_diverse}, to evaluate diverse predictions, we sample multiple candidate predictions for each historical motion, and report the best results (Best-of-Many~\cite{apratim18cvpr2}) over the candidates for each metric.

\noindent{\bf Baselines.} As our work introduces a new task, a baseline directly from prior research is not readily available. To facilitate comparisons with existing work, we adapt the following baselines from tasks of \textit{human motion generation} and \textit{object dynamic prediction}. (i) \textbf{InterVAE}: transformer-based VAEs~\cite{petrovich22temos,petrovich2021action,transformer} have been widely adopted for human motion prediction and synthesis. We employ this framework and extend it to our human-object interaction setting. (ii) \textbf{InterRNN}: we adopt a long short-term memory network (LSTM)-based~\cite{hochreiter1997long,RempeDynamics2020} predictor and enable the prediction of HOIs. For skeletal representations, we further include (iii) \textbf{CAHMP}~\cite{corona2020context} and (iv) \textbf{HO-GCN}~\cite{9714029}. As there is no publicly available codebase for the implementation, we report the results in ~\cite{corona2020context,9714029}, marked with * in Table~\ref{table:HOID}. We also implement CAHMP and present the result.
\begin{table}
\caption{\textbf{Quantitative results} on the BEHAVE dataset~\cite{bhatnagar22behave}, demonstrating the effectiveness of our diffusion model and the correction.}
\vspace{0.5em}
\label{table:BEHAVE}
\centering
\resizebox{\columnwidth}{!}{
\begin{tabular}{ccccc}
\hline\hline
\multirow{2}{*}{Method} & \multicolumn{4}{c}{BEHAVE~\cite{bhatnagar22behave}}                                       \\ \cline{2-5} 
                        & \multicolumn{1}{c}{MPJPE-H} $\downarrow$ & Trans. Err. $\downarrow$  & Rot. Err. $\downarrow$  & Pene. $\downarrow$  \\ \hline
InterRNN                &    165                         &  139           &    267        &   314      \\
InterVAE                &         145                   &    125       &   268        &  222     \\
\ours~w/o correction (\textbf{Ours})                &  \textbf{140}                           &  \textbf{123}         &    256       &  228     \\
\ours~(full) (\textbf{Ours})  &          \textbf{140}                  &     \textbf{123}        &    \textbf{226}      &  \textbf{164}     \\ \hline\hline
\end{tabular}}
\end{table}
\begin{table}
\caption{\textbf{Quantitative results} on the Human-Object Interaction dataset~\cite{9714029}. We evaluate our model in challenging scenarios with \textit{unseen instances} in the training data. The results show the effectiveness and generalizability of \ours~and the correction. * marks results directly reported from~\cite{9714029}.}
\vspace{0.5em}
\centering
\resizebox{\columnwidth}{!}{
\begin{tabular}{ccccc}
\hline\hline
Method& MPJPE-H $\downarrow$ & MPJPE-O $\downarrow$ & Trans.~Err. $\downarrow$ & Rot.~Err. $\downarrow$\\    \hline    \hline
HO-GCN*~\cite{9714029} & 111 & 153 & 123 & 303 \\
CAHMP*~\cite{corona2020context} & 107 & 167 & N/A & N/A \\
\hline
InterRNN   &124         & 127&109 & 151  \\  
CAHMP~\cite{corona2020context} & 111 & 132 & 111 & 164 \\
InterVAE   &108         & 125&100 & 178    \\
\ours~w/o Correction (\textbf{Ours})  &  \textbf{105}&117 & 92& 158       \\
\ours~w/ Correction (\textbf{Ours})  &\textbf{105} & \textbf{84}& \textbf{60}&\textbf{120}  \\
\hline\hline
\end{tabular}}
\label{table:HOID}

\end{table}
\begin{table}
\caption{\textbf{Quantitative results} on the BEHAVE dataset~\cite{bhatnagar22behave}. We generate \textit{multiple predictions} and report the lowest score across different samples. Here we focus on long-term forecasting, where we \textit{autoregressively generate 100 frames} of future interactions. Our method with interaction correction outperforms pure diffusion, and the improvement is more significant with more samples.}
\vspace{0.5em}
\label{table:BEHAVE_diverse}
\centering
\resizebox{\columnwidth}{!}{
\begin{tabular}{cccccc}
\hline\hline
\multirow{2}{*}{$\#$ of samples} & \multirow{2}{*}{\ours~(ours)} & \multicolumn{4}{c}{Best-of-Many}                                       \\ \cline{3-6} 
                        & & \multicolumn{1}{c}{MPJPE-H} $\downarrow$ & Trans. Err. $\downarrow$  & Rot. Err. $\downarrow$  & Pene. $\downarrow$  \\ \hline\hline
\multirow{2}{*}{1} & w/o correction  & 400  & 384 & 644 & 236 \\
& full  & \textbf{392} & \textbf{374} & \textbf{632} & \textbf{88}  \\ \hline
\multirow{2}{*}{2} & w/o correction  & 382 & 365 & 636 & 211 \\
& full   & \textbf{374} & \textbf{350} & \textbf{601} & \textbf{83}   \\ \hline
\multirow{2}{*}{5} & w/o correction  & 371 & 349 & 610 &  191  \\
& full  & \textbf{361} & \textbf{331} & \textbf{545} & \textbf{65} \\ \hline
\multirow{2}{*}{10} & w/o correction  & 361 & 341 & 601 & 187\\
& full & \textbf{348} & \textbf{318} & \textbf{523} & \textbf{59}\\ \hline\hline
\end{tabular}}
\end{table}
\begin{figure}
    \centering
    \includegraphics[width=\columnwidth]{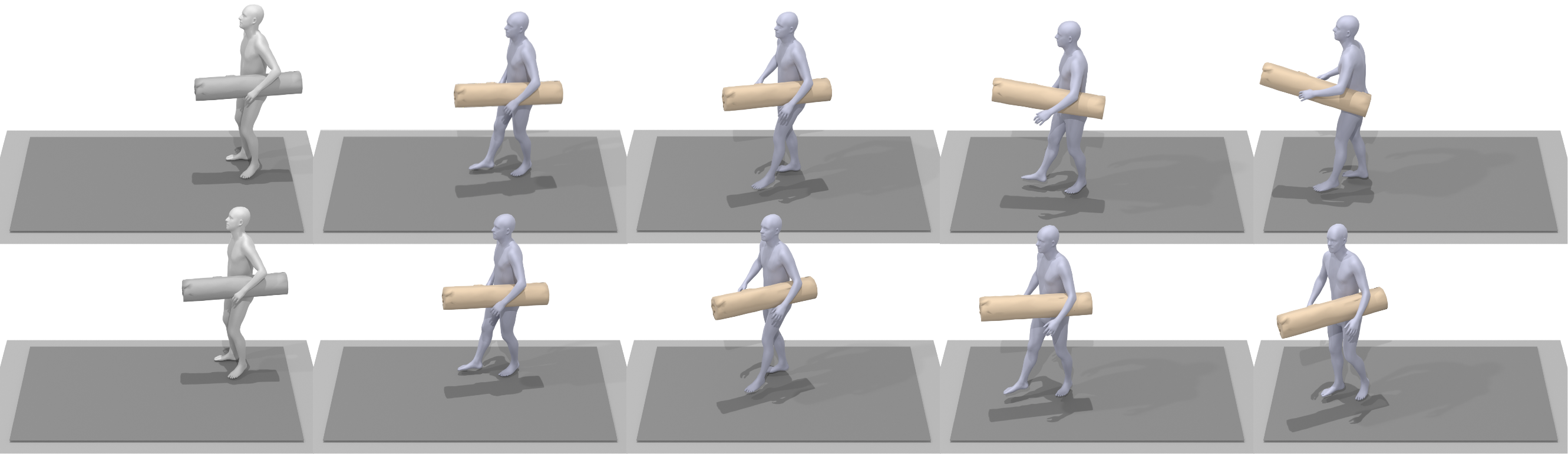}
    \caption{\textbf{Qualitative results} on the BEHAVE dataset~\cite{bhatnagar22behave}. We place two different samples of the predicted interactions. Our approach can generate diverse and legitimate predictions.}
    \label{fig:BEHAVE_diverse}
\end{figure}

\noindent\textbf{Implementation Details.}
The interaction diffusion model comprises 8 transformer~\cite{transformer} layers for the encoder and decoder, with training involving a batch size of 32, a latent dimension of 256, and 500 epochs. The interaction predictor includes 10 frequency bases for DCT/IDCT~\cite{ahmed1974discrete}, with training conducted using a batch size of 32 and 500 epochs. For the Human-Object Interaction dataset, we do not apply contact and penetration losses, since they are not applicable for skeleton representation. For autoregressive inference, we use predicted last few frames as the past motion and generate the next prediction, \etc
Additional implementation details are provided in the Supplementary.

\begin{figure}
    \centering
    \includegraphics[width=\columnwidth]{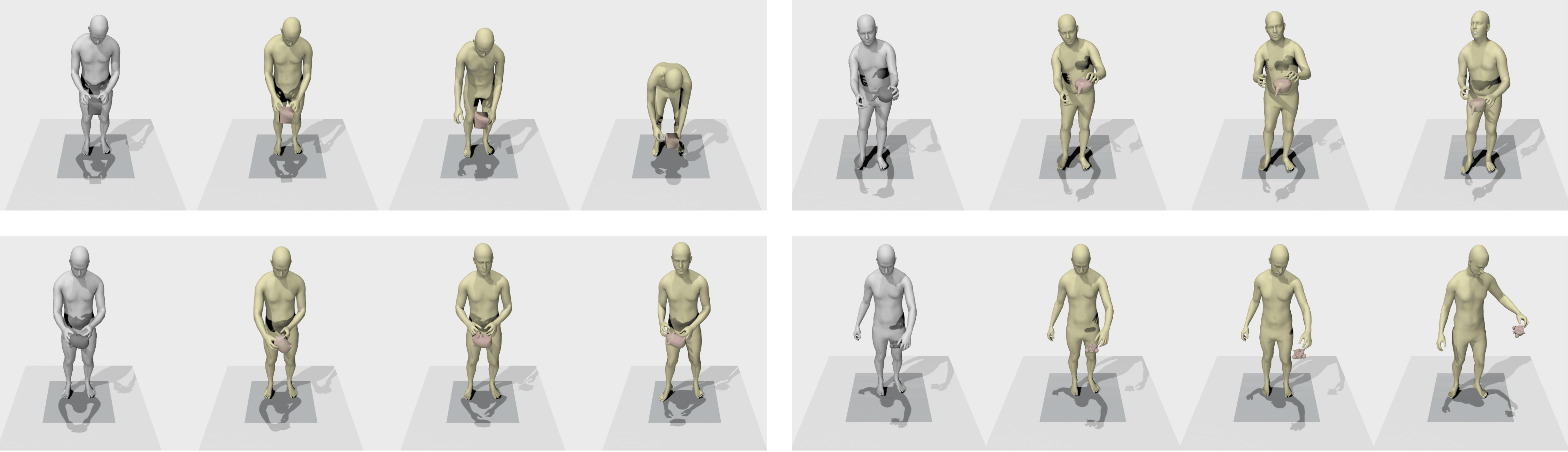}
    \caption{\textbf{Generalization} of \ours~on the GRAB dataset~\cite{taheri2020grab}. The predicted human bodies and objects are in color while the past interactions are in gray. We visualize four frames of each sequence at 0, 0.33, 0.66, and 1.0s. Our approach can directly generalize to this different dataset containing novel small-size objects.} 
    \label{fig:GRAB}
\end{figure}
\subsection{Quantitative Results}

We compare our proposed method \ours~with two baselines on the BEHAVE (Table~\ref{table:BEHAVE}) and Human-Object Interaction (Table~\ref{table:HOID}) datasets. We demonstrate the superiority of \ours~over the two baselines in all metrics for both datasets. Moreover, we observe and validate that the performance of \ours~is improved by incorporating interaction correction. Specifically, the correction step results in more plausible interactions with reduced penetration artifacts, as demonstrated in Table~\ref{table:BEHAVE}. Additionally, \ours~with interaction correction provides more precise object motions. In Table~\ref{table:BEHAVE_diverse}, following the standard Best-of-Many evaluation~\cite{apratim18cvpr2}, we demonstrate that as more predictions are sampled, the best predictions are closer to the ground truth. Note that our full method shows a significant improvement over pure diffusion with more samples and longer horizons. Furthermore, the precise object motion generated by our \ours~in turn improves the accuracy of predicted human motion even with the same interaction diffusion model, as evidenced in Table~\ref{table:BEHAVE_diverse}. The reason behind this is that by injecting more accurate object motion into the diffusion model, human motion generated by the diffusion is also positively affected and can be corrected.
\begin{table}
\caption{\textbf{User study} on the BEHAVE dataset~\cite{bhatnagar22behave}. We obtain pairwise human voting results comparing our method with baselines and alternatives introduced in Sec.~\ref{ablation}. Under human evaluation, the full model outperforms baselines regarding physical fidelity.}
\vspace{0.5em}
\label{table:BEHAVE_user_study}
\centering
\resizebox{\columnwidth}{!}{
\begin{tabular}{cccccc}
\hline\hline
\multirow{2}{*}{Model pair} & \multicolumn{3}{c}{Physical fidality}                                       \\ \cline{2-5} 
& ground truth & InterDiff (full) & w/o correction & w/o relative  \\ \hline
ground truth & N/A & 73.0\% & 69.6\% & 88.8\%   \\
InterDiff (full)  & 27.0\% & N/A & \textbf{67.8\%} & \textbf{67.8\%}  \\
w/o correction  & 30.4\% & 32.2\% & N/A & 67.2\%     \\
w/o relative  & 11.2\% & 32.2\% & 32.8\% & N/A   \\ \hline\hline
\end{tabular}}
\end{table}
\begin{figure}
    \centering
    \includegraphics[width=\columnwidth]{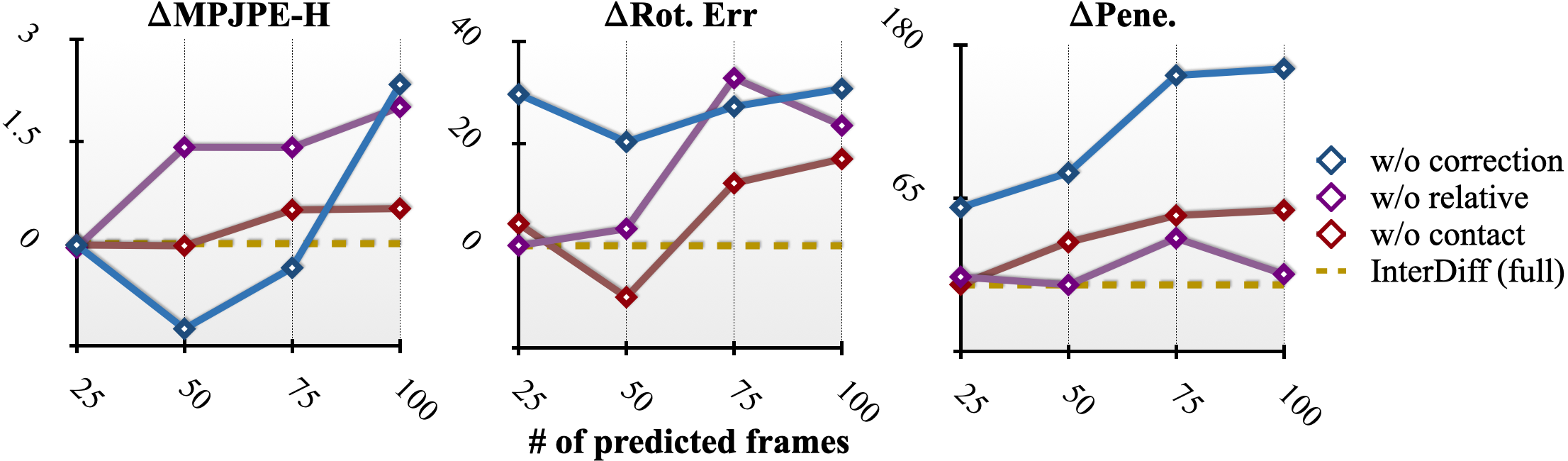}
    \caption{\textbf{Ablation study} on the BEHAVE dataset~\cite{bhatnagar22behave}. We compare our pipeline with various alternatives introduced in Sec.~\ref{ablation}. We normalize the scores of `full model' to 0. The results show the superiority of `full model' over others in the long horizon.}
    \label{fig:BEHAVE_ablation_2}
\end{figure}
\subsection{Qualitative Results}

Consistent with the quantitative results, we observe that our approach \ours~with the interaction correction yields more plausible HOI predictions than the one without interaction correction across various cases, as illustrated in Figure~\ref{fig:BEHAVE_qual}.
Furthermore, the efficacy of our method extends to the Human-Object Interaction dataset (Figure~\ref{fig:HOI_qual}), showing that our approach accommodates the skeletal representation and effectively predicts future HOIs across \textit{a diverse range of actions and unseen objects} with convincing outcomes. In Figure~\ref{fig:GRAB}, we generalize our method trained on the BEHAVE dataset to the GRAB dataset. Our approach effectively adapts to the new dataset that focuses on grasping small objects {\em without any fine-tuning}, further validating the generalizability of our approach. Figure~\ref{fig:BEHAVE_diverse} illustrates that our approach can generate diverse and legitimate HOIs. We provide additional demo videos on the project website.

To assess the motion plausibility, we conduct a double-blind user study, as shown in Table~\ref{table:BEHAVE_user_study}. We design pairwise evaluations between ground truth, \ours~(full), \ours~without interaction correction (`w/o correction'), and \ours~with the correction step yet not having coordinate transformation involved (`w/o relative'). We generate 100 frames of future interactions for comparisons. With a total of 30 pairwise comparisons, 23 human judges are asked to determine which interaction is more realistic. Our method has a success rate of $67.8\%$ against baselines.
\begin{figure*}
    \centering
    \includegraphics[width=\textwidth]{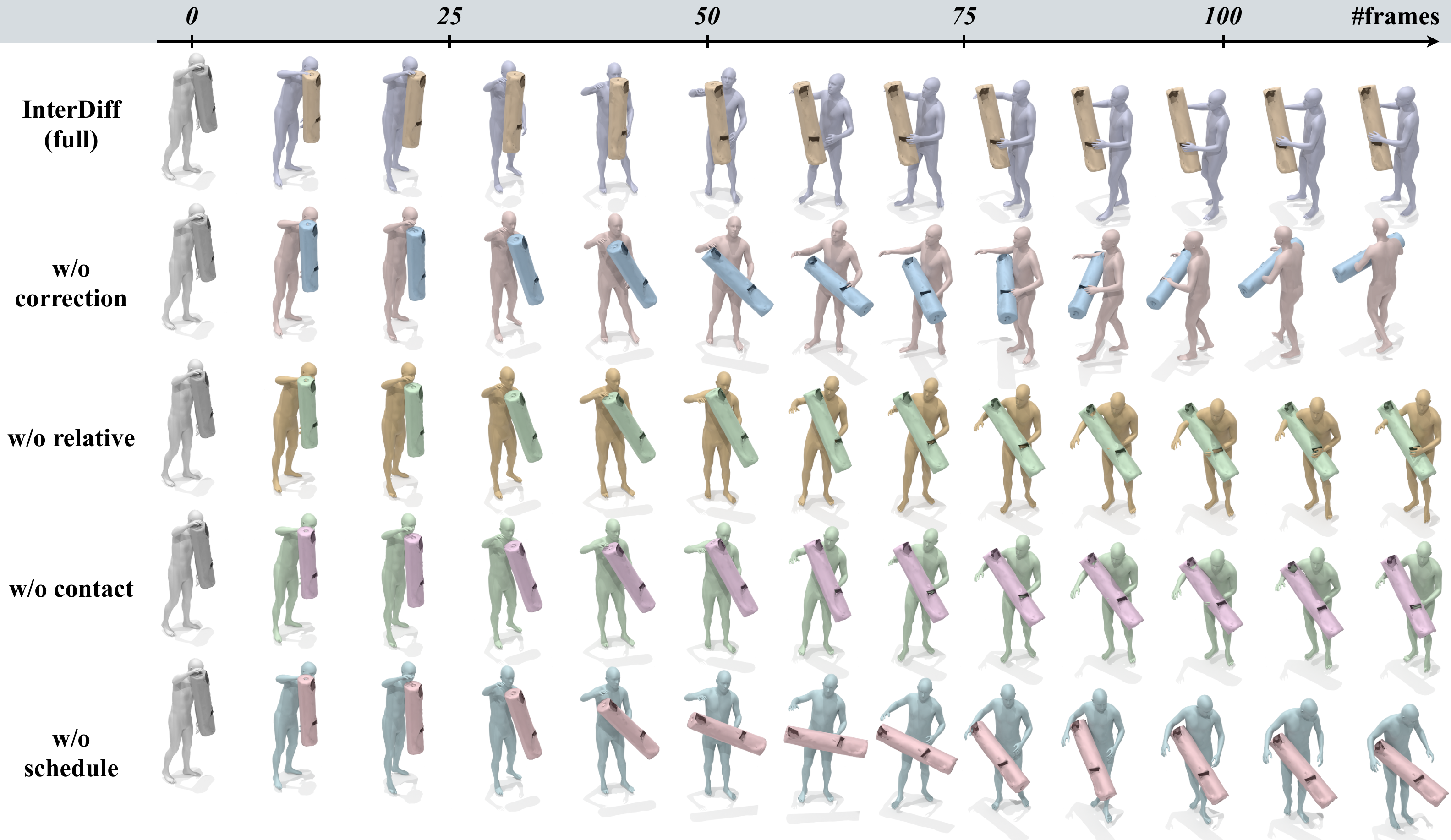}
    \caption{\textbf{Ablation study} on the BEHAVE dataset~\cite{bhatnagar22behave}. We show starting HOIs in gray and predicted HOIs sampled every 10 frames (30 FPS), up to 4 seconds. The ablated variants of our \ours~produce HOIs containing contact floating and penetration artifacts.}
    \vspace{0.5em}
    \label{fig:BEHAVE_ablation}
\end{figure*}
\subsection{Ablation Study}\label{ablation}

We conduct an ablation study (Figures~\ref{fig:BEHAVE_ablation_2} and \ref{fig:BEHAVE_ablation}) to evaluate the efficacy of different components in our proposed interaction correction (Sec.~\ref{correction}).
Figure~\ref{fig:BEHAVE_ablation} displays smooth long-term sequences generated by each ablated variant. We show that our full model, \ours, produces plausible long-term HOIs, while \ours~without the correction step results in contact floating and penetration artifacts. Then we ablate on each component inside the correction step. In the absence of reference system transformation (`w/o relative'), the object motion integrated into the diffusion fails to align with the contact's motion, resulting in significant contact penetration, especially in the long term. This highlights the critical role of the reference system in the interaction correction step. Furthermore, removing contact and penetration losses (`w/o contact') also leads to unrealistic outcomes. Finally, blindly applying correction without considering the quality of intermediate denoised results (`w/o schedule') may lead to the contact floating, as applying correction may destabilize the good quality of the original motion.
In Figure~\ref{fig:BEHAVE_ablation_2}, we further provide quantitative evidence of the effectiveness of our method, where the full model significantly outperforms the variants, especially in correcting the accumulation of errors in long-term autoregressive generation. Additional ablations are available in the Supplementary, including an evaluation of the effectiveness of DCT/IDCT in promoting simple motion patterns.
\section{Discussions}
We propose a novel task, coined as 3D human-object interaction prediction, considering the intricate real-world challenges associated with this domain.  To ensure the validity of physical interactions, we introduce an interaction diffusion framework, \ours, which effectively generates vivid interactions while simultaneously reducing common artifacts such as contact floating and penetration, with minimal additional computational cost. Our approach shows effectiveness in this novel task and thus holds significant potential for a wide range of real-world applications. A future direction would be generalizing our work to human interaction with more complex environments, such as with more than one dynamic object, more complicated objects, \eg, articulated and deformable objects, and with other humans.
\noindent \textbf{Limitations.} 
We've demonstrated that our \ours~framework is able to produce high-quality and diverse HOI predictions, without the use of post-optimization and physics simulators. Artifacts such as contact inconsistency are still observed in some generated results, though the artifacts are largely alleviated by interaction correction. Nonetheless, \ours~with correction provides effective results that can be directly applied post-optimization to improve quality. More illustrations are available on the project website.

\noindent\small{\textbf{Acknowledgement.} This work was supported in part by NSF Grant 2106825, NIFA Award 2020-67021-32799, the Jump ARCHES endowment, the NCSA Fellows program, the Illinois-Insper Partnership, and the Amazon Research Award. This work used NVIDIA GPUs at NCSA Delta through allocations CIS220014 and CIS230012 from the ACCESS program.}

\clearpage
{\small
\bibliographystyle{ieee_fullname}
\bibliography{egbib}
}

\clearpage
\appendix
\setcounter{table}{0}
\setcounter{figure}{0}
\renewcommand\thetable{\Alph{table}}
\renewcommand\thefigure{\Alph{figure}}

In this supplementary material, we include additional method details and experimental results: (1) We provide a demo video, which is explained in detail in Sec.~\ref{demo}. 
(2) We present additional information on our approach including the network architecture and learning objectives in Sec.~\ref{methodology}. (3) We provide additional implementation details in Sec.~\ref{implementation}.  (4) We show additional ablation studies in Sec.~\ref{ablation_supp}.

\section{Visualization Video}\label{demo}
In addition to the qualitative results in the main paper, we provide demos on the project website that showcase more comprehensive visualizations of the task, 3D human-object interaction (HOI) forecasting, and further demonstrate the effectiveness of our method.
In demos, we visualize that without our proposed physics-informed correction step, pure diffusion produces implausible interactions, which is consistent with the results presented in Sec.~\ref{experiments} of the main paper. In addition, we demonstrate that our method \ours~can forecast {\em diverse and extremely long-term} HOIs, while also maintaining their physical validity. Intriguingly, we observe that our method InterDiff consistently produces smooth and vivid HOIs, {\em even in cases where the ground truth data exhibit jitter patterns} from the motion capture process. Finally, we emphasize the impact and effectiveness of our contact-based coordinate system.

\section{Additional Details of Methodology}\label{methodology}

\subsection{Interaction Diffusion}
In Sec.~\ref{diffusion} of the main paper, we have highlighted our proposed \ours~pipeline. Here, we explain the architecture and the learning objectives in detail.

\noindent{\bf Architecture.}
In the reverse diffusion process, the encoder and decoder consist of several transformer layers, respectively. We set the first and last layers as the original transformer layer~\cite{attention}, while the self-attention module in the middle layers is equipped with QnA~\cite{Arar_2022_CVPR}, a local self-attention layer with learnable queries similar to~\cite{raab2023single}. 
The encoder contains an additional PointNet~\cite{qi2017pointnet} that extracts the feature of the object in the canonical pose. This shape encoding is directly added to the encoding of the past interaction, which is further processed by the transformer encoder.

\noindent{\bf Learning Objectives.}
As mentioned in the main paper, we disentangle the learning objective into rotation and translation losses for the human state $\boldsymbol h$ and the object state $\boldsymbol o$, respectively.
The original learning objective is denoted as
\begin{align}
\begin{split}
    \boldsymbol x_0(t) &= \mathcal{G}(\boldsymbol x_t, t, \boldsymbol c), \\
    \mathcal{L}_r &= \mathbb{E}_{t \sim [1, T]} \|\boldsymbol x_0(t) - \boldsymbol x\|^2_2,
\end{split}
\end{align}
where $\boldsymbol x_0(t)$ is the result given by the reverse process at step $t$, and $\boldsymbol x$ is the ground truth data, as defined in Sec.~\ref{diffusion} of the main paper.

The disentangled objectives are denoted as
\begin{align}
\begin{split}
    \mathcal{L}_h &= \mathbb{E}_{t \sim [1, T]} \|\boldsymbol h_0(t) - \boldsymbol h\|^2_2,\\
    \mathcal{L}_o &= \mathbb{E}_{t \sim [1, T]} \|\boldsymbol o_0(t) - \boldsymbol o\|^2_2,
\end{split}
\end{align}
where $\boldsymbol h_0(t), \boldsymbol h$ are the human motion generated by the diffusion model and the ground truth data, respectively. And $\boldsymbol o_0(t), \boldsymbol o$ are the denoised object motion and the ground truth, respectively.

To promote a smooth interaction over time, we introduce velocity regularizations as:
\begin{align}
\begin{split}
    \mathcal{L}_{vh} &= \mathbb{E}_{t \sim [1, T]} \|\boldsymbol h_0^{H+1:H+F}(t) - \boldsymbol h_0^{H:H+F-1}(t)\|^2_2, \\
    \mathcal{L}_{vo} &= \mathbb{E}_{t \sim [1, T]} \|\boldsymbol o_0^{H+1:H+F}(t) - \boldsymbol o_0^{H:H+F-1}(t)\|^2_2.
\end{split}
\end{align}

\subsection{Interaction Correction}
\noindent{\bf Architecture.} Here, we use SMPL~\cite{loper2015smpl}-represented human interactions as example, while we extract markers~\cite{zhang2021mojo} over the body meshes as reference. The skeleton-based interaction will follow the same process but use joints as reference. We represent the object motion under \textit{every} reference system as a spatial-temporal graph $\mathbf G^{1:H} \in \mathbb R^{H \times (1 + |\mathcal{M}|) \times D_o}$, where $D_o$ is the number of features for object poses, $1 + |\mathcal{M}|$ correspond to $1$ ground reference system and $|\mathcal{M}|$ marker-based reference systems, as mentioned in Sec.~\ref{prediction} of the main paper. Following~\cite{mao2019learning}, we first replicate the last frame $F$ times and get $\widehat{\mathbf G}^{1:H+F} \in \mathbb R^{(H+F) \times (1 + |\mathcal{M}|) \times D_o}$, then transform it into the frequency domain. Specifically, given the defined $M$ discrete cosine transform (DCT)~\cite{ahmed1974discrete} bases $\mathbf C \in \mathbb R^{M \times (H + F)}$, the graph is processed as
\begin{align}
\begin{split}
    \tilde {\mathbf G}^{1:H+F} = \mathbf C \widehat{\mathbf G}^{1:H+F}.
\end{split}
\end{align}

After applying multiple spatial-temporal graph convolutions to obtain the result $\tilde {\mathbf G'}^{1:H+F}$, we convert it back to the temporal domain, denoted as
\begin{align}
\begin{split}
    \widehat{\mathbf G'}^{1:H+F} = \mathbf C^\mathsf{T} \tilde {\mathbf G'}^{1:H+F},
\end{split}
\end{align}
where we extract the future frames $\widehat{\mathbf G'}^{H:H+F}$. As described in Sec.~\ref{prediction} of the main paper, from this graph, we index the specific future object motion with the informed reference system $s$ and then convert the motion back to the ground reference.

\noindent{\bf Learning Objectives.} Similar to the loss functions introduced for interaction diffusion, we denote two objectives as
\begin{align}
\begin{split}
    \mathcal{L}_o &= \|\widehat{\boldsymbol o}^{1:H+F} - \boldsymbol o^{1:H+F}\|^2_2,\\
    \mathcal{L}_{vo} &= \|\widehat{\boldsymbol o}^{2:H+F} - \widehat{\boldsymbol o}^{1:H+F-1}\|^2_2,
\end{split}
\end{align}
where we denote the obtained object motion including the recovered past motion as $\widehat{\boldsymbol o}^{1:H+F}$, while the ground truth object motion is $\boldsymbol o^{1:H+F}$. 
We adopt the contact loss $\mathcal{L}_c$ to encourage body vertices and object vertices close to the object surface and body surface, respectively. And the penetration loss $\mathcal{L}_p$ employs the signed distances of human meshes to penalize mutual penetration between the object and human. For more details, please refer to~\cite{wu2022saga}. Note that for skeletal representation, we do not apply $\mathcal{L}_c$ and $\mathcal{L}_p$.

\section{Additional Details of Experimental Setup}\label{implementation}

\noindent{\bf Additional Implementation Details.}
For interaction diffusion, the weight of each loss term $(\lambda_h, \lambda_o, \lambda_{vh}, \lambda_{vo}) = (1, 0.1, 0.2, 0.02)$. For interaction prediction, the weight of each loss term $(\lambda_o, \lambda_{vo}, \lambda_c, \lambda_p) = (1, 0.1, 1, 0.1)$. 
\begin{figure}
    \centering
    \includegraphics[width=\columnwidth]{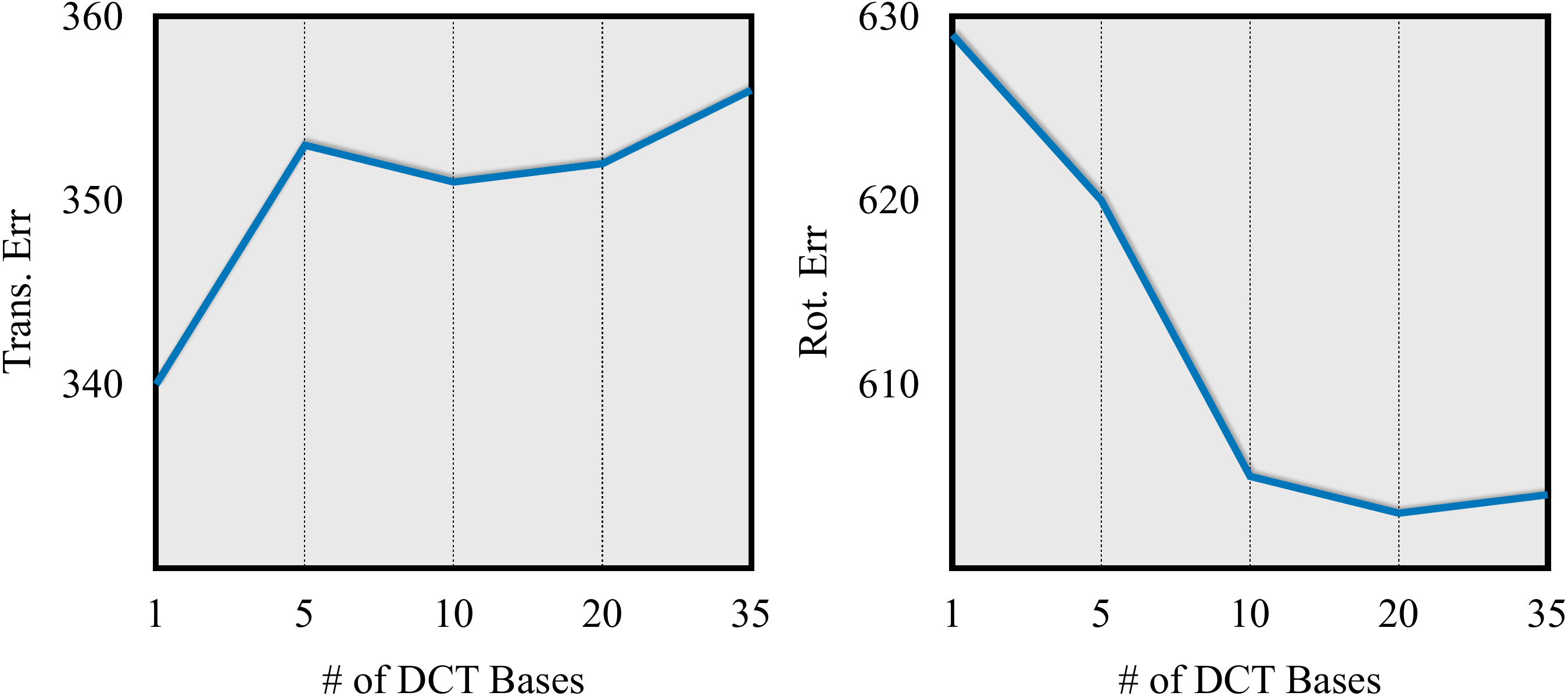}
    \caption{\textbf{Ablation study} on the BEHAVE dataset~\cite{bhatnagar22behave}. We evaluate the long-term forecasting where we autoregressively generate 100 frames of future interactions. To balance the performance in predicting rotations and translations, we set the number of DCT bases to 10.}
    \label{fig:DCT}
\end{figure}

\section{Additional Ablation Studies}\label{ablation_supp}

\noindent{\bf Effect of the number of DCT bases.} In Figure~\ref{fig:DCT}, we compare the performance when different numbers of DCT bases are used for the interaction predictor. The results show that as the number of DCT bases increases, the translation error increases, while the rotation error decreases. The reason might be that rotation is more difficult to learn and requires more parameters. However, translation relative to the reference system is very easy to model. To balance the two errors, we choose the number 10.

\end{document}